\documentclass[11pt]{article}
\usepackage{nodalida2017}
\usepackage{mathptmx}
\usepackage{url}
\usepackage{xfrac}
\usepackage{amsmath}
\usepackage{latexsym}
\usepackage{booktabs}

\title{Redefining Context Windows for Word Embedding Models: \\ An Experimental Study}

\author{Pierre Lison  \\
 Norwegian Computing Center \\
Oslo, Norway \\
{\tt plison@nr.no} \\\And
Andrei Kutuzov\\
Language Technology Group  \\
University of Oslo \\
{\tt andreku@ifi.uio.no} \\}

\date{}

\begin{document}
\maketitle

%

\begin{abstract}

Distributional semantic models learn vector representations of words through the {\it contexts} they occur in. Although the choice of context (which often takes the form of a sliding window) has a direct influence on the resulting embeddings, the exact role of this model component is still not fully understood.  This paper presents a systematic analysis of context windows based on a set of four distinct hyper-parameters. We train continuous Skip-Gram  models on two English-language corpora for various combinations of these hyper-parameters, and evaluate them on both lexical similarity and analogy tasks. Notable experimental results are the positive impact of cross-sentential contexts and the surprisingly good performance of right-context windows.

\end{abstract}

\section{Introduction}

Distributional semantic models represent words through real-valued vectors of fixed dimensions, based on the distributional properties of these words observed in large corpora. Recent approaches such as prediction-based models \cite{mikolov2013} and \textit{GloVe} \cite{pennington2014glove} have shown that it is possible to estimate dense, low-dimensional vectors (often called \textit{embeddings}) able to capture various functional or topical relations between words. These embeddings are used in a wide range of NLP tasks, including part-of-speech tagging, syntactic parsing, named entity recognition and semantic role labelling; see \cite{collobert2011natural,lin-EtAl:2015:NAACL-HLT,zhou-xu:2015:ACL-IJCNLP,lample-EtAl:2016:N16-1}, among others.

As recently shown by \cite{levy2015improving}, the empirical variations between embedding models are largely due to differences in hyper-parameters (many of which are tied to the underlying definition of context) rather than differences in the embedding algorithms themselves. In this paper, we further develop their findings with a comprehensive analysis of the role played by context window parameters when learning word embeddings. Four specific aspects are investigated:
\begin{enumerate}
  \setlength\itemsep{2pt}
\item The \textit{maximum size} of the context window;
\item The \textit{weighting scheme} of context words according to their distance to the focus word;
\item The relative \textit{position} of the context window  (symmetric, left or right side);
\item The treatment of \textit{linguistic boundaries} such as end-of-sentence markers. 
\end{enumerate}

The next section \ref{sec:background} provides a brief overview on word embeddings and context windows. Section \ref{sec:analysis} describes the experimental setup used to evaluate the influence of these four aspects.  Finally, Section \ref{sec:results} presents and discusses the results.

\section{Background}\label{sec:background}


The works of \cite{bengio2003neural} and \cite{mikolov2013} introduced a paradigm-shift for distributional semantic models with new prediction-based algorithms outperforming the existing count-based approaches \cite{baroni_dontcount}. The {\it word2vec} models from \cite{mikolov2013}, comprising the Continuous Skip-gram and the Continuous Bag-of-Words algorithms, are now a standard part of many NLP pipelines. 



Despite their differences, all types of distributional semantic models require the definition of a context for each word observed in a given corpus. 
Given a set of (word, context) pairs extracted from the corpus, vector representations of words can be derived through various estimation methods, such as predicting words given their contexts (CBOW), predicting the contexts from the words (Skip-Gram), or factorizing the log of their co-occurrence matrix (\textit{GloVe}). In all of these approaches, the choice of context is a crucial factor that directly affects the resulting vector representations. The most common method for defining this context is to rely on a \textit{window} centered around the word to estimate (often called the \textit{focus word})\footnote{Other types of context have been proposed, such as dependency-based contexts \cite{DBLP:conf/acl/LevyG14} or multilingual contexts \cite{bicompare:16}, but these are outside the scope of the present paper.}. The context window thus determines which contextual neighbours are taken into account when estimating the vector representations. 

The most prominent hyper-parameter associated to the context window is the maximum window size (i.e.~the maximum distance between the focus word and its contextual neighbours). This parameter is the easiest one to adjust using existing software, which is why it is comparatively well studied. Larger windows are known to induce embeddings that are more `topical' or `associative', improving their performance on analogy test sets, while smaller windows induce more `functional' and `synonymic' models, leading to better performance on similarity test sets \cite{goldberg2016primer}. 

However, the context window is also affected by other, less obvious hyper-parameters. Inside a given window, words that are closer to the focus word should be given more weights than more distant ones. To this end, CBOW and Continuous Skip-gram  rely on a \text{dynamic window} mechanism where the actual size of the context window is sampled uniformly from 1 to $L$, where $L$ is the maximum window size. This mechanism is equivalent to sampling each context word $w_j$ with a probability that decreases linearly with the distance $|j-i|$ to the focus word $w_i$:
\begin{align}
P(w_j|w_i) = & \sum_{window=1}^{L}P(w_j|w_i,window)  P(window)  \nonumber \\
 = & \frac{1}{L} (L-|j-i|+1) \nonumber
\end{align}
where $window$ is the actual window size (from 1 to $L$) sampled by the algorithm.  Similarly, the co-occurrence statistics used by \textit{GloVe} rely on harmonic series where words at distance $d$ from the focus word are assigned a weight $\frac{1}{d}$.  For example, with the window size 3, the context word at the position 2 will be sampled with the probability of $\sfrac{2}{3}$ in \textit{word2vec} and the probability of $\sfrac{1}{2}$ in \textit{GloVe}.

Another implicit hyper-parameter is the symmetric nature of the context window. The  \textit{word2vec} and \textit{GloVe} models pay equivalent attention to the words to the left and to the right of the focus word. However, the relative importance of left or right contexts may in principle depend on the linguistic properties of the corpus language, in particular its word ordering constraints. 

Finally, although distributional semantic models do not themselves enforce any theoretical limit on the boundaries of context windows, word embeddings are in practice often estimated on a sentence by sentence basis, thus constraining the context windows to stop at sentence boundaries. However, to the best of our knowledge, there is no systematic evaluation of how this sentence-boundary constraint affects the resulting embeddings.



\section{Experimental setup}\label{sec:analysis}

To evaluate how context windows affect the embeddings, we trained Continuous Skip-gram with Negative Sampling (SGNS) embeddings for various configurations of hyper-parameters, whose values are detailed in Table \ref{table:hyper-parameters}. In particular, the ``weighting scheme''  encodes how the context words should be weighted according to their distance with the focus word. This hyper-parameter is given two possible values: a linear weighting scheme corresponding to the default \textit{word2vec} weights, or an alternative scheme using the squared root of the distance.

\begin{table}[h]
\begin{tabular}{l|l} 
\textbf{Hyper-parameter} & \textbf{Possible values} \\ 
\midrule
Max. window size & $\{1,2,5,10\}$ \\
Weighting scheme & $\{\frac{L-d+1}{L}$,  $\frac{L-\sqrt{d}+1}{L}\}$ \\
Window position & \{left, right, symmetric\} \\
Cross-sentential & \{yes, no\} \\ 
Stop words removal & \{yes, no\} \\ 
\end{tabular}
\caption{Range of possible hyper-parameter values evaluated in the experiment.}
\label{table:hyper-parameters}
\end{table}

The embeddings were trained on two English-language corpora: Gigaword v5 \cite{en-gigaword5}, a large newswire corpus of approx.~4 billion word tokens, and the English version of OpenSubtitles \cite{opensubtitles2016}, a large repository of movie and TV subtitles, of approx.~700 million word tokens. The two corpora correspond to distinct linguistic genres, Gigaword being a corpus of news documents (average sentence length 21.7 tokens) while OpenSubtitles is a conversational corpus (average sentence length 7.3 tokens). OpenSubtitles notably contains a large number of non-sentential utterances, which are utterances lacking an overt predicate and depend on the surrounding dialogue context for their interpretation \cite{fernandez2006non}.  The corpora were lemmatized and POS-tagged with the Stanford CoreNLP \cite{manning-EtAl:2014:P14-5} and each token was replaced with its lemma and POS tag. Two versions of the corpora were used for the evaluation: one raw version with all tokens, and one filtered version after removal of stop words and punctuation. 

The word embeddings were trained with 300-dimensional vectors, 10 negative samples per word and 5 iterations. Very rare words (less than 100 occurrences in Gigaword, less than 10 in OpenSubtitles) were filtered out. 
The models were then evaluated using two standard test workflows: Spearman correlation against SimLex-999 semantic similarity dataset \cite{Hil:Rei:Kor:15} and accuracy on the semantic sections of the Google Analogies Dataset \cite{Mik:Chen:Cor:13}. 


\section{Results}\label{sec:results}
All in all, we trained 96 models on Gigaword (GW) and 96 models on OpenSubtitles  (OS)\footnote{Encompassing different values of window size, weighting scheme, window position, cross-sentential boundaries and stop-words removal ($4 \times 2 \times 3 \times 2 \times 2 = 96 $).}. Figure 1 illustrates the results for the SGNS embeddings on lexical similarity and analogy tasks using various types of context windows. The main findings from the experiments are as follows.
\subsubsection*{Window size}
As expected for a lexical similarity task \cite{schutze1993vector}, narrow context windows perform best with the SimLex999 dataset, which contains pairs of semantically similar words (not just related). For the analogy task, larger context windows are usually beneficial, but not always: the word embeddings trained on OpenSubtitles perform best with the window size of 10, while the best results on the analogy task for Gigaword are obtained with the window size of 2. 

\subsubsection*{Window position}
Table \ref{table:symmetry} shows how the position of the context window influences the average model performance. Note that symmetric windows of, for instance, 10 are in fact 2 times larger than the `left' or `right' windows of the same size, as they consider 10 words both to the left and to the right of the focus word. This is most likely why  symmetric windows consistently outperform `single-sided' ones on the analogy task, as they are able to include twice as much contextual input.

\begin{figure*}

\hspace{-7mm}\includegraphics[scale=0.44]{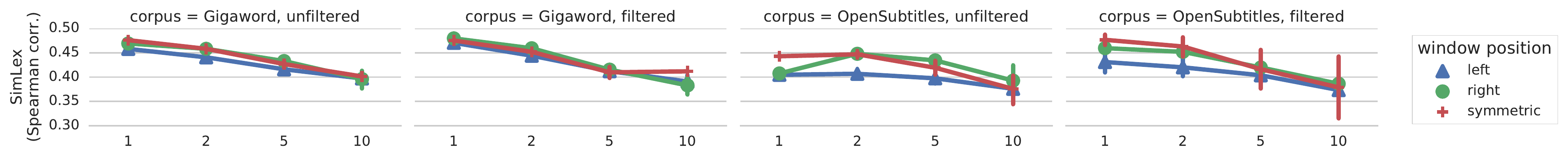} \vspace{-5mm}

\hspace{-7mm}\includegraphics[scale=0.44]{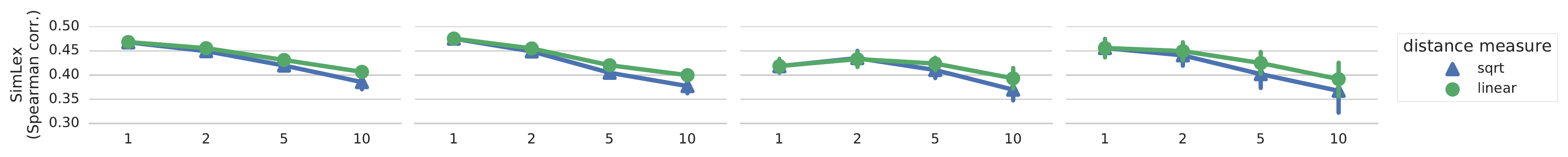} \vspace{-5mm}

\hspace{-7mm}\includegraphics[scale=0.44]{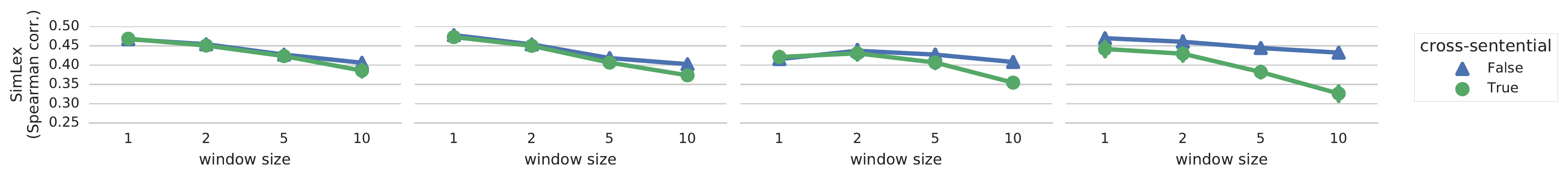}  \vspace{-2mm}

\hspace{-7mm}\includegraphics[scale=0.44]{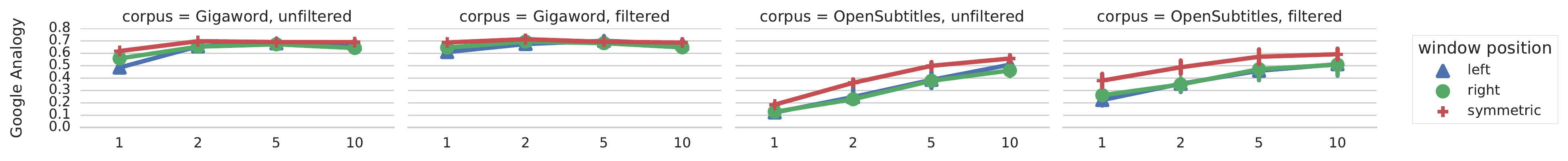} \vspace{-5mm}

\hspace{-7mm}\includegraphics[scale=0.44]{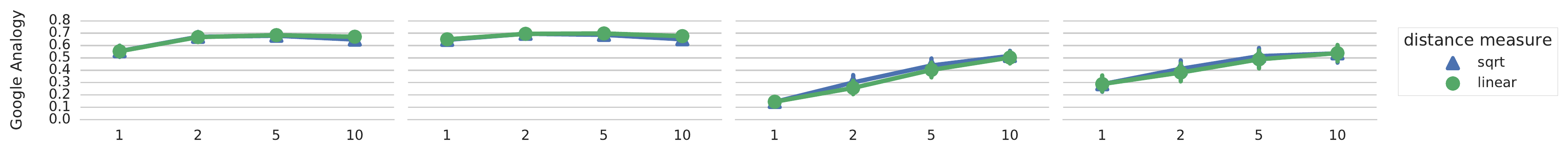} \vspace{-5mm}

\hspace{-7mm}\includegraphics[scale=0.44]{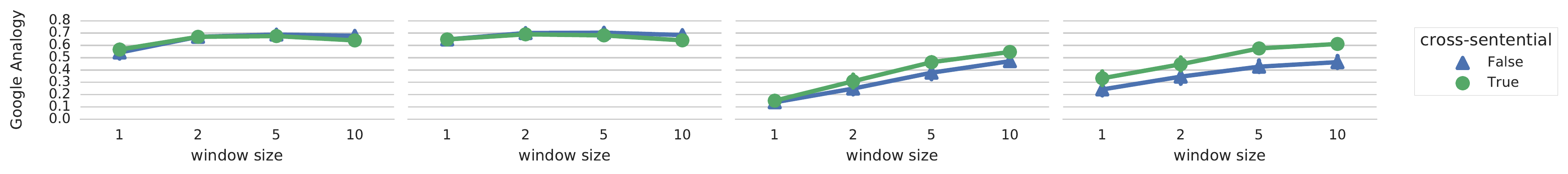} \vspace{-4mm}

\label{fig:results}
\caption{Results for the SGNS word embeddings trained with various types of context windows.}
\end{figure*}


\begin{table}
\begin{tabular}{l|c|c} 
\textbf{Window position}& \textbf{SimLex999} & \textbf{Analogies} \\
\midrule
OS left &0.40&0.35\\
OS right &0.43&0.35 \\
OS symmetric &0.43&\textbf{0.45}\\
\midrule
GW left &0.43&0.64\\
GW right &0.44&0.65\\
GW symmetric &0.45&\textbf{0.68}\\
\end{tabular}
\caption{Average performance across all models depending on the window position.}
\label{table:symmetry}
\end{table}

However, the average performance on the semantic similarity task (as indicated by the Spearman correlation with the SimLex999 test set) does not exhibit the same trend. `Left' windows are indeed worse than symmetric ones, but `right' windows are on par with the symmetric windows for OpenSubtitles and only one percent point behind them for Gigaword. It means that in many cases (at least with English texts) taking into account only $n$ context words to the right of the focus word is sufficient to achieve the same performance with SimLex999 as by using a model which additionally considers $n$ words to the left, and thus requires significantly more training time.

\subsubsection*{Cross-sentential contexts}
The utility of cross-sentential contexts depends on several covariates, most importantly the type of corpus and the nature of the evaluation task. For similarity tasks, cross-sentential contexts do not seem useful, and can even be detrimental for large window sizes. However, for analogy tasks, cross-sentential contexts lead to improved results thanks to the increased window it provides. This is especially pronounced for corpora with short sentences such as OpenSubtitles (see Table \ref{table:cross}). 

\begin{table}
\begin{tabular}{l|c|c} 
\textbf{Cross-sentential}& \textbf{SimLex999} & \textbf{Analogies} \\
\midrule
OS False &\textbf{0.44}&0.34\\
OS True  &0.40&\textbf{0.43}\\
\midrule
GW False &0.44&0.66\\
GW True &0.44&0.65\\
\end{tabular}
\caption{Average performance across all models with and without cross-sentential contexts.}
\label{table:cross}
\end{table}


\subsubsection*{Weighting scheme}
Our experimental results show that none of the two evaluated weighting schemes (with weights that decrease respectively linearly or with the square-root of the distance) gives a consistent advantage averaged across all models. However, the squared weighting scheme is substantially slower (as it increases the number of context words to consider for each focus word), decreasing the training speed about 25\% with window size 5. Thus, the original linear weighting scheme proposed in \cite{mikolov2013} should be preferred.

\subsubsection*{Stop words removal}
As shown in Table \ref{table:stop}, the removal of stop words does not really influence the average model performance for the semantic similarity task. The analogy task, however, benefits substantially from this filtering, for both corpora. Although not shown in the table, filtering stop words also significantly decreases the size of the corpus, thereby reducing the total time needed to train the word embeddings.

\begin{table}
\begin{tabular}{l|c|c} 
\textbf{Stop words removal}& \textbf{SimLex999} & \textbf{Analogies} \\
\midrule
OS no removal  &0.41&0.34\\
OS with removal &0.42&\textbf{0.43}\\
\midrule
GW no removal &0.44&0.64\\
GW with removal &0.44&\textbf{0.68}\\
\end{tabular}
\caption{Average performance across all models depending on the removal of stop words.}
\label{table:stop}
\end{table}

\section{Conclusion}
Our experiments demonstrate the importance of choosing the right type of context window when learning word embedding models. The two most prominent findings are (1) the positive role of cross-sentential contexts and (2) the fact that, at least for English corpora, right-side contexts seem to be more important than left-side contexts for similarity tasks, and achieve a performance comparable to that of symmetric windows. 

In the future, we wish to extend this study to the CBOW algorithm, to other  weighting schemes (such as the harmonic series employed by \textit{GloVe}), and to non-English corpora.

\bibliographystyle{acl}
\bibliography{nodalida2017}

\end{document}